\documentclass[
]{ceurart}

\usepackage{listings}
\usepackage{tcolorbox}
\usepackage{amsmath}
\usepackage{xcolor}
\usepackage{framed}

\colorlet{shadecolor}{gray!10}

\begin{document}

\copyrightyear{2024}
\copyrightclause{Copyright for this paper by its authors.  Use permitted under Creative Commons License Attribution 4.0  International (CC BY 4.0).}

\conference{AIEB 2024: Workshop on Implementing AI Ethics through a Behavioural Lens | co-located with ECAI 2024, Santiago de Compostela, Spain}

\title{Framework for developing and evaluating ethical collaboration between expert and machine}

\author{Ayan Banerjee}[
orcid=0000-0001-6529-1644,
email=abanerj3@asu.edu,]

\address{School of Computing and Augmented Intelligence,
  Arizona State University, Tempe, 85281, USA}

\author{Payal Kamboj}[%
orcid=0000-0003-1950-3444,
email=pkamboj@asu.edu,
]

\author{Sandeep Gupta}[%
orcid=0000-0002-6108-5584,
email=sandeep.gupta@asu.edu,
]

\begin{abstract}
  Precision medicine is a promising approach for accessible disease diagnosis and personalized intervention planning in high-mortality diseases such as coronary artery disease (CAD), drug-resistant epilepsy (DRE), and chronic illnesses like Type 1 diabetes (T1D). By leveraging artificial intelligence (AI), precision medicine tailors diagnosis and treatment solutions to individual patients by explicitly modeling variance in pathophysiology. However, the adoption of AI in medical applications faces significant challenges, including poor generalizability across centers, demographics, and comorbidities, limited explainability in clinical terms, and a lack of trust in ethical decision-making. This paper proposes a framework to develop and ethically evaluate expert-guided multi-modal AI, addressing these challenges in AI integration within precision medicine. We illustrate this framework with case study on insulin management for T1D. To ensure ethical considerations and clinician engagement, we adopt a co-design approach where AI serves an assistive role, with final diagnoses or treatment plans emerging from collaboration between clinicians and AI.
\end{abstract}

\begin{keywords}
  Deep Learning \sep
  Multimodal \sep
  Ethical \sep
  Human-machine collaboration
\end{keywords}

\maketitle

\section{Introduction}


Precision medicine offers promising avenues for diagnosing and personalizing treatment plans for high-mortality diseases like coronary artery disease~\cite{b33}, drug-resistant epilepsy~\cite{b18}~\cite{frontiersKamboj}~\cite{Banerjee2023}, and chronic illnesses such as Type 1 diabetes~\cite{b5}. Precision medicine can help tailoring diagnoses and treatments to individual patients by accounting for variations in pathophysiology with the help of Artificial Intelligene (AI). However, AI's integration in medical practice faces challenges such as poor generalizability across different centers and populations due to change in scanners, age, gender; limited clinical explainability; and ethical concerns. We present a framework for developing and ethically evaluating expert-guided multi-modal AI to address these challenges by using a co-design approach, where AI functions in an assistive role, with final diagnosis or treatment plan resulting from a collaboration between clinicians and AI.  Multi-modal AI (MAI) improves information content in rare pathological samples by integrating knowledge from various modalities. Our approach further augments information content by incorporating expert knowledge from clinicians. We integrate deep learning (DL) models trained with multi-modal data that learns pathophysiological trends across the patient population, with expert guided first principles based mechanistic models or digital twins, trained with patient-specific data to capture personalized clinical presentations of disease. This integration is designed to optimize generalized performance, while maintaining the capacity to map AI outputs back to clinically relevant parameters of the digital twin, thus enabling the generation of explanations. 

\begin{figure*}[t]
\centering
\includegraphics[width=\textwidth]{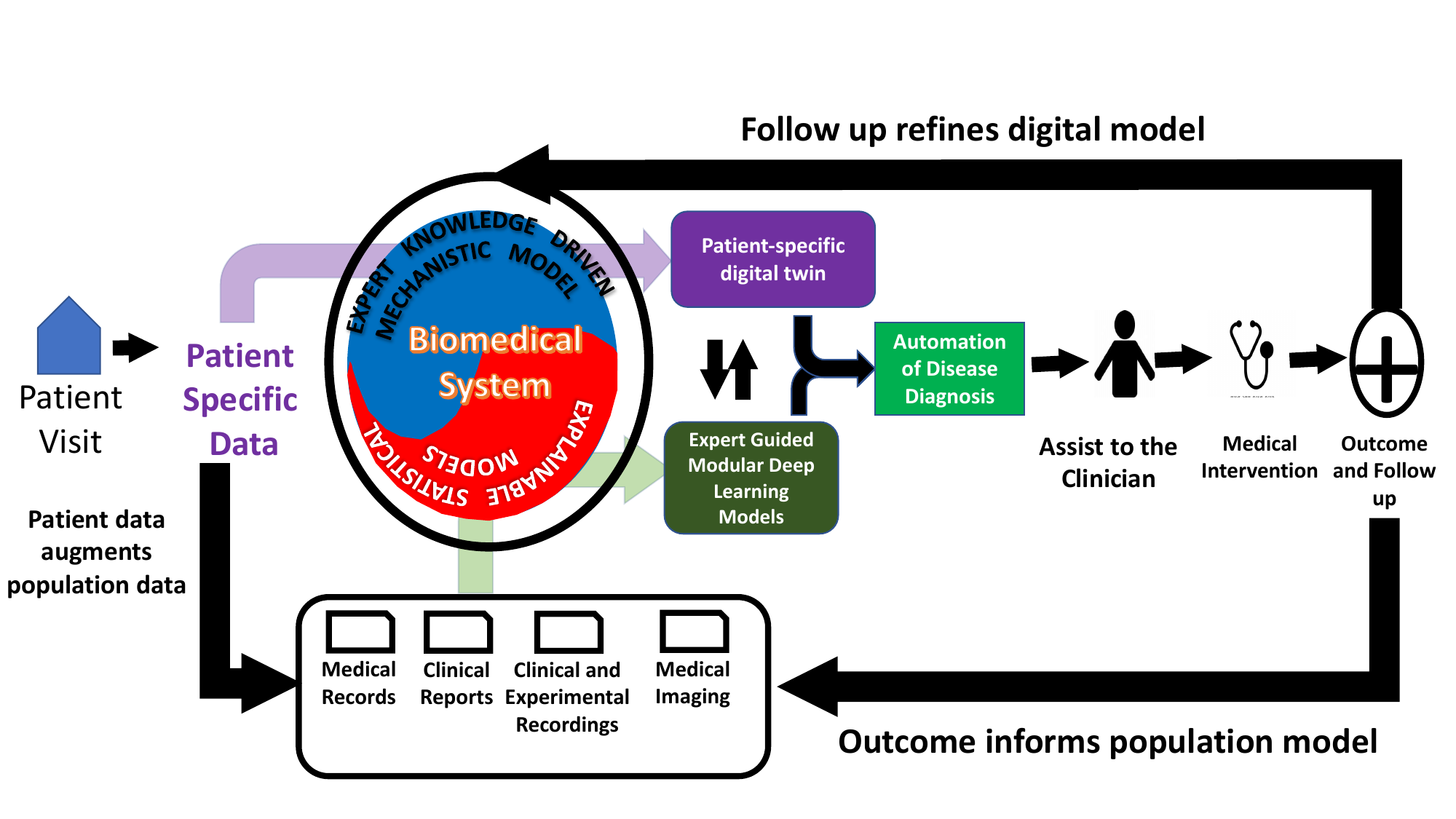}
\caption{Figure 1: Precision medicine approach with integration of digital twin and multimodal AI guided by expert knowledge.}
\label{fig:7}
\end{figure*}
In the endocrine system (T1D insulin management) challenge, large language models (LLM) trained with multi-modal T1D data can be integrated with Bergman minimal model-based patient-specific simulators, trained on continuous glucose monitor and insulin data. This will derive time schedules of automated insulin delivery (AID) system configuration settings and meal / bolus insulin doses to optimize glycemic control during exercise, pregnancy or aging. In the nervous system challenge, convolutional neural networks trained with resting state fMRI images can be combined with propositional logic-based digital twin trained with Blood Oxygen Level Dependent time / frequency features to detect seizure onset zones (SOZs) from rs-fMRI in DRE patients. In the disease prediction challenge, transformer models trained on ESE images can be integrated with generative models fitted to patient-specific baseline ECGs, to provide sex-unbiased CAD prediction. 

We consider three stages of AI technology lifecycle including: conceptualization, development and calibration. In AID, in the conceptualization phase, we use expert knowledge of glucose-insulin dynamics to develop patient specific simulators. In the development phase, we use the simulators fitted with data to analyze plans generated by LLMs, and in the calibration phase we evaluate the LLM-generated plans for various rare scenarios related to exercise, pregnancy and aging in T1D (beneficence + non-maleficence + autonomy). In SOZ detection, we can incorporate age-independent expert knowledge during the conceptualization stage, use multicenter training data for both children and adults in the development phase, and conduct age-stratified, cross-center evaluations during the calibration phase to develop unbiased SOZ detectors with for adults and children (justice). In CAD detection, we can incorporate differences in clinical presentation of CAD during the conceptualization phase, we can use gender specific digital twins as input to the transformer based CAD prediction model during the development phase, and in the calibration phase, we can use attention maps to determine if the transformer model effectively utilizes gender differences to improve positive predictive value (PPV) for women (justice). In this paper, we will show the case study of insulin management for T1D.

\subsection{Biomedical System Level Challenge} 
Accessible disease diagnosis and personalized intervention planning are crucial for improving health outcomes for high-mortality diseases such as ischemic heart disease~\cite{b1}, and drug resistant epilepsy ~\cite{b2,b3,b4} as well as chronic illnesses such as Type 1 diabetes (T1D)~\cite{b5}. In such domains, precision medicine~\cite{b6} has become increasingly important. As shown in Figure 1, precision medicine moves away from a one-size-fits-all approach to disease management, advocating for personalized strategies that explicitly address individual variances. At the core of this approach is a patient-specific digital twin, configured to mimic the relevant pathophysiological characteristics of an individual. For the digital twin to be effective, it must accurately represent pathophysiological data and explain variations in terms of clinically relevant factors. Hence, most typically a first principle based mechanistic model of physiology is chosen as the base model of the digital twin which is then fitted to the patient data using machine learning (ML). A high-fidelity mechanistic model consists of various clinically relevant variables describing the pathophysiology of the patient and hence digital twins are commonly learned using multi-modal data. The digital twin can be used to synthesize pathophysiological data under various conditions, which is then utilized by an automation software to aid in disease diagnosis, prediction of disease prognosis, and generation of treatment plans. To maintain bioethical standards, in this framework, we take a co-design approach where engineers interface with stakeholders such as clinicians and patients at every stage starting from conceptualization, where expert knowledge guides initial MAI based precision medicine solution design, development, where clinicians and bioethics experts determine the datasets, metrics, and provide feedback on generalizability and explainability of the implemented solution, and calibration, where clinicians and bioethics experts provide feedback on in-the-field usage of the solution. As such the MAI based automation software is designed as an assistive technology to the team of clinicians who are responsible for making the final decision in patient health management. Post consultation with the clinician, the patient level outcomes and follow-up tests generate more multi-modal data that iteratively updates the digital twin. 

Automated insulin delivery (AID) systems are FDA approved Class II medical device used for insulin management in individuals aged 2 years or older with Type 1 Diabetes (T1D) to keep blood glucose level within a target range of 70 mg/dL to 180 mg/dL, also known as Time in Range (TIR), for at least 70\% of the day~\cite{b7}. However, FDA approved AID systems currently struggle to effectively manage day to day activities such as exercise or life events such as pregnancy. The biomedical challenge is to develop usage plans for existing FDA approved AID systems that consider specific glycemic variation characteristics due to exercise or pregnancy. A usage plan consists of a time aligned sequence of AID configuration settings such as glucose target, insulin sensitivity factor (ISF), carbohydrate ratio (CR), and user actions such as meal intake and bolus insulin dosage. A patient specific digital twin that models the endocrine system based on mechanistic models such as Bergman Minimal Model and mimics patient specific variabilities due to exercise or pregnancy can be used as a simulator to test various usage plans to improve AID performance. Since usage plans for mitigating glycemic variability in exercise and pregnancy are not well established, novel treatment plans need to be explored. A large language model (LLM) trained with multi-modal data including time series continuous glucose monitor and insulin pump data, TIR plots and doctor’s notes in publicly available reports of patients with similar pathophysiology, expert opinions on novel usage plans, can effectively search novel usage plans. These plans can then be evaluated using the patient specific digital twin. Comparative analysis in terms of clinically relevant factors can be used by clinicians to advice the user on the optimal usage plan that can improve TIR while avoiding severe hypoglycemia.

As seen from the aforementioned application domain, it is evident that multi-modal data driven learning, a significant subclass of AI, serves as the foundational technology for precision healthcare. However, the primary roadblock for precision healthcare is related to fundamental drawbacks of using AI in life critical medical applications including: 
a) \textbf{Lack of generalized performance:} AI techniques often struggle with distribution shifts, which occur when data from a different center, demographics, or comorbidity profile are used during deployment.  
b) \textbf{Lack of transparency and explainability in AI driven decision making:} While black box AI models may have good performance, their decisions often lack explainability in terms of relevant clinical factors. On the other hand, an explainable machine may not always achieve good generalized performance.  
c) \textbf{Lack of trust in AI’s capability of ethical decision making:} the pillars of bioethics, beneficence, non-maleficence, patient autonomy, and justice are not always preserved in black box models~\cite{kamzin2020evaluating}. 
In this paper, we present a framework for developing and analyzing AI techniques that are generalizable, explainable, ethically trustworthy and can be integrated as the core component of precision healthcare to address the biomedical system level challenges covering unbiased diagnosis, personalized treatment planning, and accessible per-surgical evaluation. 

\section{Method}

The main hypothesis in our solution to AI driven precision medicine is that integration of expert knowledge acquired by clinicians in-the-field with data driven AI can enable generalized, transparent, explainable, and ethical automation. The core philosophy behind this hypothesis is that expert knowledge is gained by clinicians through years of field experience across centers and are familiar with disease manifestations across demographics~\cite{b18}. Consequently, this knowledge acquired by experts in the field is likely to be generalizable across centers, demographics and co-morbidity profiles. Moreover, it is highly likely that expert clinician knowledge will highlight the clinical factors that are relevant for a disease profile. Hence, if an AI technique can utilize expert knowledge in a modular fashion, then its decisions can be explained in terms of relative sensitivity of the output to each knowledge module. Furthermore, relevant bioethical questions are often a function of the expert clinician’s experience. Hence, integration of expert knowledge in all three stages of AI technology lifecycle including: conceptualization, development, and calibration~\cite{b16} can potentially result in generalizable, explainable, and ethical AI for precision healthcare. 

Characteristics of expert knowledge: Expert knowledge can be of many types including knowledge about:
a)	Data distribution across centers, which depend on measurement devices, changes in test protocols ~\cite{b17},  
b)	Patient population, including differences in demography and co-morbidity profiles across centers and geographical locations~\cite{b18},
c)	performance metrics, that are relevant for different patient population~\cite{b19},
d)	clinically relevant features, which are used for diagnosis and determining optimal therapeutic plans ~\cite{b12} ~\cite{b18},
e)	problem domain knowledge in terms of rules satisfied by clinical representation of pathophysiology in multi-modal data~\cite{b21}~\cite{modality_selection2024}~\cite{CAD_2024}, and
f)	bioethics, including factors affecting beneficence, non-maleficence, justice, and patient autonomy~\cite{b16}.

Expert knowledge on data distribution, and patient population has been used previously in unimodal AI under the umbrella of domain knowledge guided AI ~\cite{b22}~\cite{kamboj2023merging}, and knowledge of performance metrics and bioethics is most typically utilized in any ethical AI method ~\cite{b23}. Explainable AI approaches utilize clinically relevant features to explain the AI outcome to the clinician ~\cite{b18}. However, problem domain knowledge remains underutilized primary due to the following fundamental properties:
a)	\textbf{Propositional logic:} Problem domain knowledge is often obtained through textual description as propositions that connects data with outcome. Their objective evaluation and incorporation in MAI require integration of symbolic AI strategies with DL. Such integration has only recently been explored in medical applications, as in our prior work~\cite{b18}, and remains untested in large scale studies. 
b)	\textbf{Multi-modality:} Problem domain knowledge highlights salient features of outcomes irrespective of data modality and hence most commonly objective evaluation of problem domain knowledge requires multi-modal data.
In addition, problem domain knowledge also suffers from two key disadvantages:
a)	\textbf{Vagueness:} Expressed in natural language, expert knowledge can be vague and subjective resulting in ambiguity in its objective evaluation. 
b)	\textbf{Conflict:} Machine interpretation of natural language expression of expert knowledge may result in conflicting propositions potentially reducing discriminative power. 

\begin{figure*}[t]
\centering
\includegraphics[width=\textwidth, clip=true,trim=0 0 0 0]{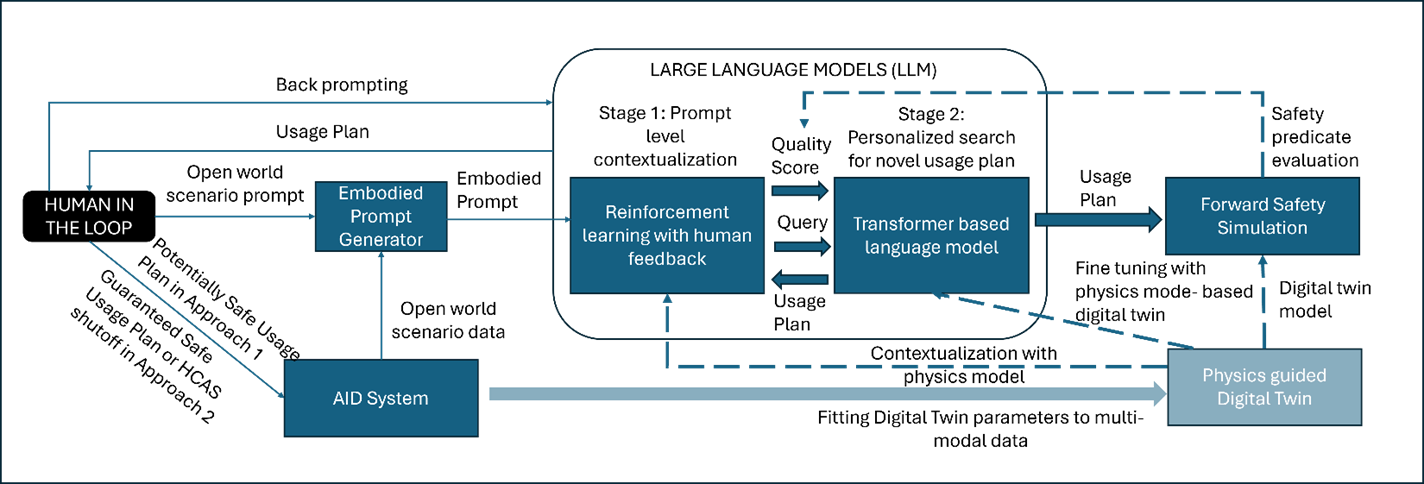}
\caption{. LLM planner for the endocrine system modeling challenge: LLMs finetuned with multi-modal data for generating safe usage plans under dynamically changing user contexts.}
\label{fig:2}
\end{figure*}
\subsection{Why multi-modal AI integrated with expert knowledge may improve generalizability, explainability and bioethics? }

In multi-modal AI (MAI), a pathophysiological response \(z\) is expressed as a parametric function of data from multiple modalities \(\{x_1, \dots, x_n\}\), \(z = f(x_1, \dots, x_n, \Theta)\), where \(\Theta\) is a set of patient-specific parameters. The aim is to derive the function \(f\), from samples of \(\{x_1, \dots, x_n, z\}\) of labelled data. Unimodal AI refers to the special case of the problem when \(n=1\). Multi-modal learning gives better generalized performance than unimodal learning only when two conditions are satisfied~\cite{b24}:

\begin{itemize}
    \item \textbf{Heterogeneity}, which requires that each modality \(x_i\) contributes some unique features important for the pathophysiology of the disease that cannot be derived from any other modality \(x_j, j \neq i\), and
    \item \textbf{Connection}, which entails that there is a function \(g_{ij}(x_i, x_j, \Theta_k)\), that relates \(x_i\) to \(x_j\) based on a subset of patient-specific parameters \(\Theta_k \subset \Theta\) and \(g_{ij}\) is a simpler function than \(f\), in terms of VC dimension \cite{b24}.
\end{itemize}

If these two conditions hold, then instead of directly learning \(f\), MAI can first utilize the modalities to learn the simpler functions \(g_{ij}\) and parameter subsets \(\Theta_k\), and then with complete knowledge of \(\Theta\), utilize the modalities to learn the function \(f\). If \(g_{ij}\) is simpler than \(f\) by \(O(N^m)\), \(m>0\), where \(N\) is the number of training samples, then MAI gives better generalized performance in learning \(f\) than unimodal AI.

The fundamental drawback in MAI theory is its assumption that the MAI learning technique has knowledge about the structure of \(g_{ij}\). One could argue that if a learner is unaware of the structure of \(g_{ij}\), then it can use a learning function \(l_{ij}\), that is not only of higher complexity than \(g_{ij}\), but it might be even more complex than directly learning \(f\), rendering MAI useless. In the proposed framework, we address this drawback using problem domain knowledge from clinicians about the structure of the pathophysiological properties of a disease.

Problem domain knowledge helps in two aspects:
\begin{itemize}
    \item \textbf{Identifying modalities} \(\{x_1, \dots, x_n\}\) such that the heterogeneity requirement is maintained, and
    \item \textbf{Identifying learning functions} \(l_{ij}\) that are of the similar structure as \(g_{ij}\).
\end{itemize}

Utilizing such expert knowledge, our solutions follow a general approach consisting of two steps:
\begin{enumerate}
    \item \textbf{Digital Twin Learning} - use a series of expert identified learning functions \(l_{ij}\) to fit multi-modal training data and derive a set of clinically relevant patient-specific parameters \(\Theta\). The parameter set \(\Theta\) along with the parameterized learning function \(l_{ij}\) acts as a digital twin that accurately models the disease-specific pathophysiology of the patient.
    \item \textbf{Expert Guided DL} – use the fitted learning functions \(l_{ij}\) and patient-specific parameter set \(\Theta\) as inputs to DL techniques to learn the function \(f\).
\end{enumerate}

Apart from providing generalized performance, the problem domain knowledge guided approach also enables explainability. This is because in the MAI approach towards learning \(f\), we first learn \(g_{ij}\) and \(\Theta\) using the structure of \(g_{ij}\) based on clinical knowledge about the pathophysiological properties. Hence, the learning functions \(l_{ij}\) and patient-specific parameters \(\Theta_k\) directly correspond to clinically relevant factors. As such, any characteristics of the final learned function \(f\) can be explained in terms of the clinically relevant factors.

In this paper, we evaluate this fundamental hypothesis that expert knowledge integration with MAI can lead to generalizable, explainable, and ethical AI.

\subsubsection{Personalization of Automated Insulin Delivery for individuals with T1D}
\textbf{Problem Statement} To develop a guardian angel software that guides users with Type 1 Diabetes (T1D) to create personalized usage plans for FDA-approved Automated Insulin Delivery (AID) systems. This software will incorporate clinician advice to ensure safe insulin delivery in response to dynamic context changes, such as daily activities (e.g., exercise) or life events/stages (e.g., pregnancy, aging). The output will be an embodied Large Language Model (LLM) interface capable of generating tailored usage plans for various glycemic control scenarios.

An embodied LLM is a pretrained language model that is retrained with real world application specific multi-modal data. Publicly available multi-modal databases including images of glucose charts, glucose insulin time series data, doctor notes can be transformed using signal specific encoders to match the dimension of the latent space in LLMs. A pretrained LLM can be incrementally retrained using multi-modal data, potentially giving it the capability to convert high level user enquiries about diabetes management to low level control configuration update.
In Figure. 2 the embodied LLM acts as a mediator between the human user and the AID and can potentially be configured to: a) model human intervention plans, b) ensure safety of the intervention plans, and c) provide assistive feedback to guide the human user towards safe personalization of the AID. The starting point to the LLM is an initial prompt that is converted to an initial plan by the LLM. Consider the following prompt:

\begin{tcolorbox}[colback=gray!10!white, colframe=gray!40!black, title=Prompt]
\textbf{I have Type 1 Diabetes and I am using the Medtronic 770G automated insulin delivery system. I want to run for 30 mins in the next hour. My current CGM reading is 85 mg/dL. My Insulin sensitivity factor setting is 50, and my carbohydrate to insulin ratio is 0.36. What set point do I set to avoid hypoglycemia?}
\end{tcolorbox}

\begin{tcolorbox}[colback=gray!10!white, colframe=gray!40!black, title=Response from GPT-3.5]
\textbf{Before Exercise:} Since your current blood sugar level is 85 mg/dL, you may want to consume a small carbohydrate-containing snack (such as fruit or a granola bar) to raise your blood sugar to a safer level before you start running. Aim to bring your blood sugar to a level between 140-180 mg/dL (7.8-10 mmol/L) before beginning your exercise.

\textbf{During Exercise:} Monitor your blood sugar levels closely using your CGM throughout your run. Aim to maintain your blood sugar within the target range of 140-180 mg/dL (7.8-10 mmol/L) by adjusting your carbohydrate intake and insulin delivery as needed. If your blood sugar starts to drop below 140 mg/dL (7.8 mmol/L), consider consuming fast-acting carbohydrates (such as glucose tablets or sports drinks) to raise your blood sugar and prevent hypoglycemia.

\textbf{After Exercise:} Continue monitoring your blood sugar levels after you finish your run, as exercise can sometimes lead to delayed hypoglycemia. Aim to keep your blood sugar within the target range of 140-180 mg/dL (7.8-10 mmol/L) during the post-exercise period. Adjust your insulin doses and carbohydrate intake for your next meal or snack based on your blood sugar readings and any trends you observe.
\end{tcolorbox}

The initial plan generated by the LLM may not be safe for the unique physiology of the user. To ensure that a plan generated by the LLM is safe for an individual the approach shown in Figure. 2 will be used. This approach will check plan safety utilizing a forward safety simulation module and iteratively modify an initial plan into a safe plan by providing the plan quality evaluation as feedback to the LLM (Figure. 2). 
\begin{figure*}[t]
\centering
\includegraphics[width=0.9\textwidth, clip=true,trim=0 0 0 0]{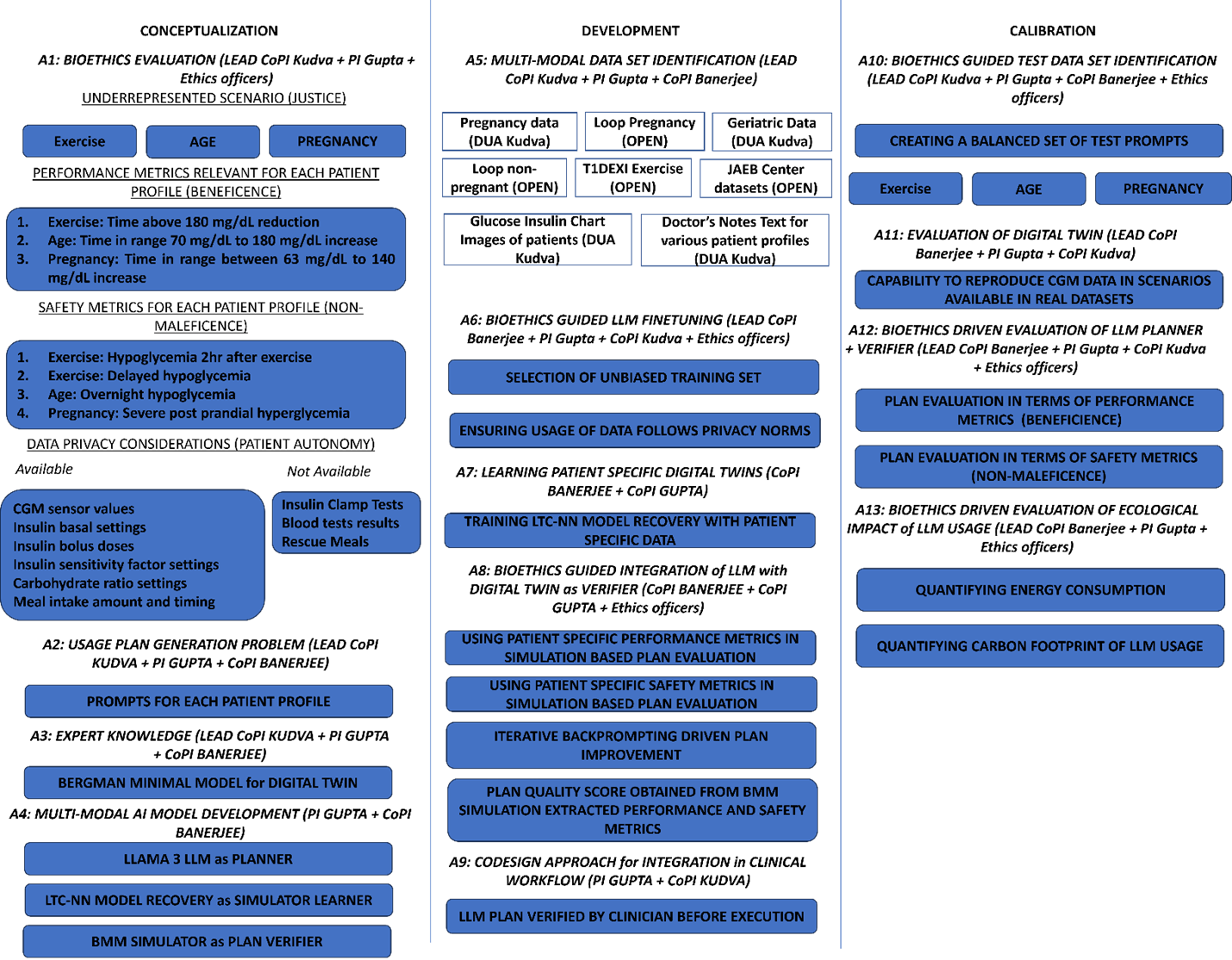}
\caption{Task Organization for the Automated Insulin Delivery biomedical challenge. This project is a collaboration between ASU and Mayo clinic, Rochester.}
\label{fig:3}
\end{figure*}

\textbf{Problem domain knowledge:} The forward safety simulator is a digital twin that will be learned from data obtained from the AID system used. Expert knowledge about the endocrine system will be utilized to define the model structure of the digital twin, such as the Bergman Minimal Model (BMM). The BMM expresses the CGM data as a function of insulin input \(f\), using a differential equation-based combination \(g_{ij}\) of insulin input, plasma insulin, interstitial insulin, and plasma glucose levels. The model parameters will be fitted to the data. To accomplish this deliverable, we undertake the research tasks discussed in Figure 3, divided into stages of ethical MAI design~\cite{b16}.

\textbf{Conceptualization:} The first task (A1) is bioethics evaluation of the problem of personalization of AID. A collaboration with a clinical centre personnel will be established to understand the beneficence, non-maleficence, and patient autonomy issues for each under-represented glycemic control scenarios such as exercise or for life events/ stages such as pregnancy and aging (patient distribution knowledge). It includes determination of performance and safety metrics, and create a set of data available to the embodied LLM architecture while maintaining the patient privacy factors (data distribution knowledge).
The second task (A2) is to identify glycemic control problems for exercise, pregnancy and aging and convert them into prompt inputs for the LLM with expert clinician guidance. Task A3 is a collaboration between the clinician and the engineering team to utilize problem domain knowledge on endocrine system modeling and determine the base structure of the digital twin. Task A4 is the multi-modal AI model development task, and it has three subtasks: 
a) \textbf{Identify the LLM to be used as the initial planner:} Two requirements of selection are that the LLM should have a fine-tunable API and it should be relatively computationally efficient. This restricts our selection to LLAMA 2/3 ~\cite{b25} or Phi 2 ~\cite{b26} as these models are much smaller in scale than enterprise level LLMs.
b) \textbf{Develop a method to fit the BMM based digital twin model structure} to the real data obtained from specific patients who are using the AID system. For this purpose, we will extend  prior work that proposes a liquid time constant neural network (LTC-NN) model recovery method that learns digital twin-based simulators for pregnant individuals with T1D~\cite{banerjee2024emily}. We apply this technique to model exercise related variabilities and fit simulators to data collected across age ranges.
c) \textbf{Implement integration of LLM planner with the digital twin based simulation driven plan safety evaluation}. We will extend our prior work that integrated LLAMA 2 with the LTC-NN driven digital twin to evaluate a single bolus insulin delivery action~\cite{b27} to now generate usage plans.

The key advantages of using multi-modal AI such as LLMs like Llama-2/3~\cite{b25} and Phi 2~\cite{b26} are:
1. \textbf{Generalization:} LLMs can find the appropriate patient cluster for an individual based on their demographic and comorbidity profile, and explore a significantly large set of applicable usage plans documented in textual forms. For example, doctor’s notes contain information on the safe usage of an AID system for T1D. LLMs can tap into such resources to derive novel plans applicable for a given user.
2. \textbf{Explainability:} Interaction with LLMs is intuitive for the AID user and may provide inherent explainability and reasoning for the generated plans. 
3. \textbf{Bioethics:} LLMs can be contextualized for understanding bioethical factors for a given geographic space and demographic cluster. Most recently, LLMs have entered the domain of bioethics and have been evaluated to be adequate in addressing several bioethical issues~\cite{b28}. Although state-of-the-art LLMs may not be able to resolve certain special cases of real world ethical dilemmas, they have been shown to be capable of at least identifying complex ethical issues that may affect a deployed AI system.

\textbf{Development:} The first task in the development phase A5 is collaboration with clinician to determine the set of multi-modal data to be used in the fine-tuning of the embodied LLM architecture. This includes whole day continuous glucose monitor (CGM) charts (images), CGM data and insulin pump data obtained for different patient population available throughout clinical studies conducted by Mayo Clicic, Rochester or open data available through JAEB Center~\cite{b29}, NIH library, or T1DEXI exercise studies~\cite{b30}. 
The next task (A6) is the finetuning of the embodied LLM architecture with the multi-modal data. The first step is to contextualize the RLHF module of the LLM with the BMM model or digital twin of the individual. This contextualization process will be done by generating domain specific prompts regarding the BMM based digital twin. The LLM will then be fine-tuned using embodied instruction prompts that encode the relationship between CGM insulin data and model parameters of the digital twin. The fine-tuned LLM is capable of correlating model parameters with operational scenario data and incorporates such causal relations in its plan search mechanism.
The task A7 is the fitting of the digital twin to real data so it can be used as a high-fidelity forward safety simulator~\cite{Banerjee2023_HILHIP}. This involves training the LTC-NN architecture with real world data related to exercise pregnancy and aging.
The next task (A8) is the bioethics guided integration of the multi-modal LLM and the digital twin. Instantiated with the usage plan derived by the LLM, a forward simulation of the recovered model is an evaluation of the safety criteria. Here patient autonomy will be considered by choosing different safety metrics that are relevant to the specific patient profile. Metrics such as robustness of signal temporal logic formula can be used to quantify the safety and performance requirements ~\cite{b27}. This safety evaluation is then passed to the reinforcement learning with human feedback (RLHF) module to modulate the plan quality score. If a plan is unsafe, then a heavy penalty is imposed in the plan quality score. This quality score can be used for further plan refinement either through the RLHF architecture or through explicit back-prompting from the user.
The task A9 is to develop an interface such that the LLM generated plan can be verified by the clinician before use fostering a co-design approach rather than a full autonomous method.

Advantages of incorporating multimodal data and expert knowledge in development phase:
1. \textbf{Generalizability:} LLMs are trained with large scale multi-modal data gathered from various sources. Hence, inherently they provide generalized results. The problem is with contextualizing and fine tuning the generalized result and making it applicable for the unique user contexts. 
2. \textbf{Explainability:} LLMs through their reasoning capacity and natural language interface is inherently explainable as evident in the sample response. Combined with chain of thought reasoning, LLMs can potentially explain a usage plan step by step to a user. 
3. \textbf{Bioethics:} With the multi-modal training the LLMs are at an inherent advantage to provide generalized performance across age, gender, disease comorbidities and other demographics. For example, an LLM fine tuned to provide safe usage plans for an individual woman can also provide expert guided plans for managing insulin delivery during pregnancy. In the example below, the Prompt 1 can be augmented to make prompt 2 with GPT 3.5 response
\begin{tcolorbox}[colback=gray!10!white, colframe=gray!40!black, title=Prompt]
\textbf{I am in the sixth week of pregnancy. What should be my meal plan throughout the day and exercise plan to maintain > 70\% time in range?}
\end{tcolorbox}

\begin{tcolorbox}[colback=gray!10!white, colframe=gray!40!black, title=Response from GPT-3.5]
Congratulations on your pregnancy! Here's a suggested meal plan and exercise plan to help you maintain greater than 70\% time in range.

\textbf{Meal Plan:}
\begin{itemize}
    \item \textbf{Breakfast (8:00 AM):} Whole grain toast with avocado. Aim for 30-45 g of carbohydrates.
    \item \textbf{Mid-Morning Snack (10:30 AM):} Greek yogurt with nuts. Aim for around 15-30 grams of carbohydrates.
    \item \textbf{Lunch (12:30 PM):} Grilled chicken salad with quinoa. Aim for 45-60 grams of carbohydrates.
    \item \textbf{Afternoon Snack (3:00 PM):} Carrot sticks with hummus. Aim for around 15-30 grams of carbohydrates.
    \item \textbf{Dinner (6:30 PM):} Baked salmon with roasted sweet potatoes. Aim for approximately 45-60 grams of carbohydrates.
    \item \textbf{Evening Snack (9:00 PM):} Cottage cheese with sliced fruit. Aim for around 15-30 grams of carbohydrates.
\end{itemize}

\textbf{Exercise Plan:}
\begin{itemize}
    \item \textbf{Morning Exercise (8:30 AM):} Walking, prenatal yoga, or swimming for 30 minutes of moderate-intensity exercise.
    \item \textbf{Afternoon Exercise (3:30 PM):} Leisurely walk or stretching routine.
    \item \textbf{Evening Relaxation (7:30 PM):} Deep breathing, meditation, or other stress relief activities before bedtime.
\end{itemize}
\end{tcolorbox}

Note that AID systems are not approved for usage in pregnancy, but the LLM is able to change recommendation-based stage of pregnancy utilizing the expert knowledge that it has gathered through multi-modal data driven training. The question remains that whether this plan is safe or not which can be assured using digital twin integration method shown in Figure 2.

\textbf{Calibration:} The first task (A10) is to identify test data following bioethical standards enforced by collaboration between the ASU and Mayo teams. The next task A11 is to test the accuracy of the digital twin in replicating the test data. Hallucination and continuous learning can be significant roadblocks to reliable plan recommendation from LLMs (A12). As such beyond plan safety and feasibility, hallucination and plan robustness need to be addressed explicitly. Plan safety can be evaluated in silico using FDA approved simulator such as the UVA PADOVA simulator. For specific user conditions such as pregnancy, the digital twin can be used to simulate scenarios and evaluate plan safety. Each plan can be used to instantiate a simulation run and forward safety can be accessed over the plan horizon. Plan feasibility can be assessed by establishing feasibility constraints based on user preferences or extracting constraints by analyzing the properties of the digital twin. Hallucination evaluation should be performed manually by expert users in the field of AID with number of irrelevant responses per 100 queries from the LLM architecture.

The advantages of the multi-modal LLM training architecture in calibration:
1. \textbf{Generalizability:} The capability of back-prompting is powerful in addressing safety, hallucination, and feasibility in an individual context. Hence, even if the LLM driven architecture is not generalized, it can be updated quickly through back-prompting based contextualization aided by the digital twin driven forward simulation-based plan quality evaluator.
2. \textbf{Explainability:} Plan quality score and the output of the digital twin based forward simulator can provide explanation as to why a plan may or may not work. It can also provide with a plan risk score that can warn the user with the risks of hypo or hyperglycemia if the plan is executed. These risks can be explained in terms of popular glycemic metrics such as projected time in range, mean glucose or hypoglycemic events. 
3. \textbf{Bioethics:} The configurable LLM architecture can be used to evaluate plans based on metrics that are traditionally not reported for AID systems, such as projected number of severe hypoglycemia events, time above range, exercise driven hypoglycemia. This enables patient autonomy through the LLM interface. One of the major impact of using LLM is the ecological footprint of the usage plan recommendation system (Task A13). LLMs are large models and executing them in the cloud cause significant power consumption and computational resource utilization. One major issue addressed at this stage is the use of distilled models that a smaller scale than LLM but still give good performance in a limited context. The smaller scale LLM can be potentially distilled to provide safe plans only for the context of exercise. We can obtain lower scale models that can be implemented in embedded computing devices.

\subsection{Ethical statement}
Recovery of updated digital has several ethical issues that need to be addressed: \textit{a) Privacy preserved digital twin recovery} - The recovery process should not require additional sensing of physiological variables. This may lead to unmeasurable state variables of the first-principles based model. As a result the problem of digitl twin recovery requires us to derive parameters for implicit dynamics, which is inherently a more difficult problem than traditional system identification. \textit{b) Ethical data collection for recovery} - The digital twin recovery process should not require the participant to undergo additional testing protocols that are not part of the normal usage of the biomedical system. Data from normal usage of the system may be insufficient for identifiability of all the parameters.
\section{Conclusions}
In this paper, we have highlighted the need for infusion of ethics in every step of development of biomedical applications. We have provided a framework and also shown a case study on ethics guided biomedical application development. This is intended as an initial template that can get the discussion started on ethics guided development and can be modified through collaboration and future discussion.

\section{Acknowledgement}
This work was supported by the Helmsley Charitable Trust under grant number 2-SRA-2017-503-M-B and by the National Science Foundation (NSF) under grant number FDTBioTech 2436801.

\bibliography{ecai.bib}

\appendix

\end{document}